\documentclass[a4paper,twoside]{article}

\usepackage{epsfig}
\usepackage{subcaption}
\usepackage{calc}
\usepackage{amssymb}
\usepackage{amstext}
\usepackage{amsmath}
\usepackage{amsthm}
\usepackage{multicol}
\usepackage{pslatex}
\usepackage{apalike}
\usepackage{algorithm2e}
\usepackage[bottom]{footmisc}
\newcommand{\x}{\textbf{x}}
\newcommand{\z}{\textbf{z}}
\DeclareMathOperator*{\pr}{Pr} % thin space, limits underneath in displays
\newcommand{\RNum}[1]{(\expandafter{\romannumeral #1\relax})}

\usepackage{amsmath}  % Needed for \DeclareMathOperator
\DeclareMathOperator*{\argmax}{arg\,max}  % The asterisk ensures the limits are placed underneath in display style

\usepackage{booktabs}
\usepackage{paralist}
\usepackage{url}

\usepackage[acronym,nomain,nonumberlist]{glossaries}
\usepackage{amsfonts}
\usepackage{wrapfig}

\usepackage{SCITEPRESS}     % Please add other packages that you may need BEFORE the SCITEPRESS.sty package.

\setacronymstyle{long-short}
\newacronym{pbpm}{PBPM}{Predictive Business Process Monitoring}
\newacronym{nap}{NAP}{Next Activity Prediction}
\newacronym{ai}{AI}{Artificial Intelligence}
\newacronym{xai}{XAI}{eXplainable Artificial Intelligence}
\newacronym{ml}{ML}{Machine Learning}
\newacronym{lstm}{LSTM}{Long Short-Term Memory}
\newacronym{dl}{DL}{Deep Learning}
\newacronym{svm}{SVM}{Support Vector Machine}
\newacronym{rnn}{RNN}{Recurrent Neural Network}
\newacronym{kpi}{KPI}{Key Performance Indicator}
\newacronym{bpm}{BPM}{Business Process Management}
\newacronym{pm}{PM}{Process Mining}
\newacronym{dt}{DT}{Decision Tree}
\newacronym{sst}{SST}{Sufficient Subset Training}
\newacronym{it}{IT}{Information Technology}
\newacronym{eos}{EOS}{End-of-Sequence}
\newacronym{ce}{CE}{Cross Entropy}
\newacronym{mae}{MAE}{Mean Absolute Error}
\newacronym{ood}{OOD}{Out-of-Distribution}

\begin{document}

\title{Self-Explaining Neural Networks for Business Process Monitoring}

% \author{\authorname{First Author Name\sup{1}\orcidAuthor{0000-0000-0000-0000}, Second Author Name\sup{1}\orcidAuthor{0000-0000-0000-0000} and Third Author Name\sup{2}\orcidAuthor{0000-0000-0000-0000}}
% \affiliation{\sup{1}Institute of Problem Solving, XYZ University, My Street, MyTown, MyCountry}
% \affiliation{\sup{2}Department of Computing, Main University, MySecondTown, MyCountry}
% \email{\{first\_author, second\_author\}@ips.xyz.edu, third\_author@dc.mu.edu}
% }

% Anonymize:
% To undo, comment in lines 69-75 AND the footnote in line 98 (and comment out lines 77-80)

\author{
\authorname{Shahaf Bassan$^*$\sup{1,2}, Shlomit Gur$^*$\sup{2}\orcidAuthor{0000-0001-5174-3689}, Sergey Zeltyn$^*$\sup{2}\orcidAuthor{0000-0003-2540-1604}, Konstantinos Mavrogiorgos$^{**}$\sup{3}\orcidAuthor{0009-0006-5534-2683}, Ron Eliav\sup{1,4} and Dimosthenis Kyriazis\sup{3}\orcidAuthor{0000-0001-7019-7214}}
\affiliation{\sup{1}School of Computer Science and Engineering, Hebrew University of Jerusalem, Jerusalem, Israel}
\affiliation{\sup{2}IBM Research, Haifa, Israel}
\affiliation{\sup{3}Department of Digital Systems, University of Piraeus, Piraeus, Greece}
\affiliation{\sup{4}Department of Computer Science, Bar-Ilan University, Giv'atayim, Israel}
% \email{shahaf.bassan@mail.huji.ac.il, shlomit.gur@ibm.com, sergeyz@il.ibm.com, komav@unipi.gr, dimos@unipi.gr}
\email{komav@unipi.gr}
}

% \author{
% \authorname{Anonymous authors\sup{1}}
% \affiliation{\sup{1}Paper under double-blind review}
% }


\abstract{Tasks in \acrfull{pbpm}, such as \acrlong{nap}, focus on generating useful business predictions from historical case logs. 
Recently, \acrlong{dl} methods, particularly sequence-to-sequence models like \acrfull{lstm}, have become a dominant approach for tackling these tasks. 
However, to enhance model transparency, build trust in the predictions, and gain a deeper understanding of business processes, it is crucial to explain the decisions made by these models. 
Existing explainability methods for \acrshort{pbpm} decisions are typically \emph{post-hoc}, meaning they provide explanations only after the model has been trained. 
Unfortunately, these post-hoc approaches have shown to face various challenges, including lack of faithfulness, high computational costs and a significant sensitivity to out-of-distribution samples. 
In this work, we introduce, to the best of our knowledge, the first \emph{self-explaining neural network} architecture for predictive process monitoring. 
Our framework trains an \acrshort{lstm} model that not only provides predictions but also outputs a concise explanation for each prediction, while adapting the optimization objective to improve the reliability of the explanation. 
We first demonstrate that incorporating explainability into the training process does not hurt model performance, and in some cases, actually improves it. 
Additionally, we show that our method outperforms post-hoc approaches in terms of both the faithfulness of the generated explanations and substantial improvements in efficiency.}

\onecolumn \maketitle \normalsize \setcounter{footnote}{0} \vfill

\def\thefootnote{*}\footnotetext{These authors contributed equally to this work}\def\thefootnote{\arabic{footnote}}
\def\thefootnote{**}\footnotetext{Corresponding author}\def\thefootnote{\arabic{footnote}}

%%%%%%%%%%%%%%%%%%%%%%%%%%%%%%%%%%%%%%%%%%%%%%%%%%%%%%%%%%%%%%%%%%%%%%%%

\section{Introduction}

\label{sec:introduction}

 % Introduction to PBPM: history, types of tasks, why important
 % Relation to prescriptive

\acrfull{pbpm} plays a crucial role in \acrfull{bpm}. 
It focuses on predicting key process metrics and outcomes, such as \acrfull{nap}, the time to process completion, and the final state of the process. 
This predictive capability is essential for identifying potential issues and bottlenecks within processes, thereby enabling preemptive measures and optimizing resource allocation. 
As a result, many \acrshort{bpm} and \acrfull{pm} software solutions now integrate predictive features into their frameworks~\cite{galanti2020explainable}. 
Typically, these prediction models use historical event logs as training data, and the input features for a specific prediction are extracted from the trace prefix of a process instance, for which the predictions are made~\cite{neu2021prediction}.

Historically, \acrshort{pbpm} utilized conventional Machine Learning (\acrshort{ml}) techniques, such as Support Vector Machines (SVMs), K-Nearest Neighbor (KNNs), and Decision Trees (DTs)~\cite{marquez2017survey}. 
Typically, these methods perform well with numeric and categorical attributes but struggle to
incorporate sequential data seamlessly.
With the advent of \acrfull{dl}, \acrfull{lstm} networks emerged as the predominant approach to \acrshort{pbpm} for several years, with one of the earliest applications of \acrshort{lstm} to \acrshort{nap} presented in $2017$~\cite{evermann2017DL}.
The field has since evolved to include other \acrshort{dl} methods, like transformers~\cite{ramamaneiro2023benchmark}.
For additional details, there are comprehensive reviews of \acrshort{dl} applications to \acrshort{pbpm}~\cite{neu2021prediction,ramamaneiro2023benchmark,ramamaneiro2024graph}.
Given that contemporary \acrshort{dl} models may have billions of parameters, the latest \acrshort{pbpm} models have become highly sophisticated and are often challenging to interpret. 
Yet, providing explanations for these predictions in a format that end-users can understand is as critical as the accuracy of the predictions themselves, especially in high-stakes environments such as healthcare and banking~\cite{jia2022role,marques2022delivering}. 
For instance, explaining why an algorithm predicts a lengthy wait for a hospital procedure or foresees a bank customer declining a loan offer is vital for offering valuable insights to stakeholders and clients.

\acrfull{xai} methods aim to address the aforementioned challenge~\cite{meske2022explainable,adadi2018peeking,arya2019one}. 
The traditional works in \acrshort{xai} primarily examine \emph{post-hoc} techniques that focus on explaining the decisions of a trained model \emph{after} its training. 
Among these, \emph{local} post-hoc explanations (in contrast to global explanations) provide insights into specific decisions made by \acrshort{ml} models for specific data points. 
Common local post-hoc \acrshort{xai} methods include additive feature attribution techniques such as LIME~\cite{ribeiro2016should} and SHAP~\cite{lundberg2017unified}, assigning an attribution score to each feature. 
A significant drawback of these methods is their implicit assumption of near-linear behavior of the model around the analyzed input~\cite{yeh2019fidelity}, which may not hold for highly non-linear contemporary \acrshort{dl} models.

%To address this limitation, an alternative approach has emerged alongside additive feature attribution techniques, which will be the main focus of this work, encompassing methods such as Anchors~\cite{ribeiro2018anchors} and SIS~\cite{carter2019made}. These methods aim to provide a different type of post-hoc explanation—specifically, obtaining a \emph{sufficient explanation}, i.e., a subset of input features that, on their own, ensure the prediction remains unchanged, regardless of the values assigned to the remaining features.

To overcome this limitation, an alternative approach has emerged alongside additive feature attribution techniques. 
This approach is the primary focus of this work and includes methods such as Anchors~\cite{ribeiro2018anchors} and SIS~\cite{carter2019made}, which aim to offer a distinct form of post-hoc explanation --- specifically, identifying a \emph{sufficient explanation}. 
This refers to a subset of input features that, by themselves, guarantee the prediction remains unchanged, regardless of the values assigned to the remaining features. 
It is well established in the literature that \emph{smaller} sufficient subsets lead to more interpretable explanations~\cite{ribeiro2018anchors,carter2019made,bassan2023towards,ignatiev2019abduction,bassan2023formally}. Consequently, methods like Anchors and SIS strive to identify a subset that is both sufficient and \emph{minimal} --- capturing the most concise set of input features that solely determine the prediction.

The literature has proposed various types of sufficient explanations, including those that provide different robustness guarantees~\cite{wu2024verix,marques2022delivering,bassan2023towards,la2021guaranteed,darwiche2020reasons,wu2024marabou}, probabilistic assurances~\cite{waldchen2021computational,arenas2022computing}, or those that use a specific baseline for evaluating the sufficiency of the subset in question~\cite{chockler2021explanations}. 
While many of these methodologies are model-agnostic, meaning they can be applied across different models, some are designed specifically for particular models like \acrshortpl{dt}~\cite{izza2020explaining}, \acrshort{dt} ensembles~\cite{ignatiev2022using,audemard2022trading}, or Neural Networks~\cite{wu2024verix,bassan2023towards,la2021guaranteed}.

%To cope with this drawback, following these feature attribution techniques, another line of works, which includes methods such as Anchors \cite{ribeiro2018anchors}, SIS~\cite{carter2019made}, attempt to generate a different form of prevelanet (post-hoc) explanation --- a subset of input features that are by themselves sufficient to determine that the prediction remains the same, regardless of the assignments to the complementary features of that subset. It is widely acknowledged in the literature that \emph{smaller} subsets offer better interpretations~\cite{ribeiro2018anchors,carter2019made,bassan2023towards,ignatiev2019abduction}, and therefore, methods that seek to obtain such subsets (e.g., Anchors and SIS) usually attempt to obtain a subset that is both sufficient and minimal --- identifying a concise subset of input features that are important and by themselves determine the prediction.

% **
% taking classical work and adding explainability in training
Although post-hoc explanation methods, including both additive attribution and sufficiency-based approaches, offer valuable insights, they encounter several problems, such as lack of faithfulness~\cite{rudin2019stop,slack2020fooling,camburu2019can}, scalability issues due to high computational demands~\cite{barcelo2020model,bassan2024local,waldchen2021computational,marzouk2025computational,marzouk2024tractability}, and sensitivity to sampling 
\acrfull{ood} assignments~\cite{hase2021out,amir2024hard}. 
To address these limitations, recent exciting developments have introduced \emph{self-explaining neural networks} (SENNs)~\cite{alvarez2018towards}, which inherently generate explanations as part of their output, potentially alleviating many of these challenges. 
% This approach has shown effectiveness in the context of additive attribution explanations~\cite{alvarez2018towards} and, more recently, sufficiency-based explanations \cite{bassan2025iclr}.
% To tackle these issues a new line of research has suggested \emph{self explaining neural networks} (SENNs)~\cite{alvarez2018towards} as a general framework under which one can train a model that inherently provides explanations as part of its input, and thus mitigating some of these prior problems. 

SENNs inherently provide \emph{additive feature attributions}, assigning importance weights to individual or high-level features. 
For instance, describing model outputs by comparing them to relevant “prototypes”~\cite{chen2019looks,wang2021self,jiang2024protogate,keswani2022proto2proto} or deriving concept coefficients through feature transformations~\cite{alvarez2018towards}. 
Other methods focus on feature-specific Neural Networks or use classifiers applied to text snippets for NLP explanations~\cite{agarwal2021neural,jain2020learning} or predict outcomes based on high-level concepts~\cite{koh2020concept,kim2023probabilistic,espinosa2022concept}. 
Finally, a recent work proposes a self-explaining architecture that inherently generates sufficient explanations for its predictions~\cite{bassan2025iclr}.

\textbf{Our contributions.} 
In this work, we present, to the best of our knowledge, the first \emph{self-explaining} neural network architecture for \acrshort{pbpm} tasks. 
To demonstrate our approach, we tackled the well-known \acrshort{nap} problem~\cite{polato2018time}. 

We built upon a widely-adopted \acrshort{nap} \acrshort{lstm}-based model~\cite{tax2017lstm}, modifying its open source code to adapt it to a self-explaining framework. 
Our proposed architecture goes beyond making predictions by also producing a second output, consisting of a concise and sufficient explanation for the prediction. 
It follows a methodology akin to the general self-explaining framework~\cite{alvarez2018towards} and its more recent adaptation to the sufficient explanation setting~\cite{bassan2025iclr}. 
However, our setting presents unique challenges, requiring the model to capture both the seq-to-seq nature of \acrshort{nap} tasks and integrate \acrshort{bpm}-specific considerations into its architecture. 

We assessed our approach using four event logs from the banking and \acrfull{it} industries. 
We conducted a comprehensive comparison between the explanations produced by our architecture and those generated by the widely used post-hoc method, Anchors~\cite{ribeiro2018anchors}, which addresses the same task of obtaining sufficient explanations, but without the self-explaining intervention. 
Our findings highlight a notable improvement in our explanations compared to the post-hoc setting. 
This includes a substantially increased faithfulness of our explanations (i.e., the proportion of explanations that were indeed sufficient) and a substantial reduction in computation time. 
Thus, we regard these findings as compelling evidence, supporting the use of self-explaining methods in the context of \acrshort{pbpm} tasks to produce more dependable and trustworthy explanations for their decisions.

The rest of the paper is organized as follows: Section~\ref{sec:prelimanries} presents background information, covering our setting, sufficient explanations, and \acrshortpl{lstm}. Section~\ref{sec:method} details our methodology. 
Section~\ref{sec:experiments} discusses the data used, the experiments conducted, and the results obtained. 
Finally, Section~\ref{sec:discussion} explores the implications of our findings and suggests avenues for future research.

\section{Preliminaries}
\label{sec:prelimanries}

\subsection{Explainability Setting}

Since our primary focus is on explaining \acrshort{nap} tasks, we can generalize this as an explanation for \emph{classification} tasks. 
Specifically, while the model we seek to explain, $f$, produces multiple types of outputs, including regression outputs, the component of interest for our explanations is fundamentally a classification output (the \acrshort{nap} part). 
We can hence formally define, without loss of generality, our interpreted model as $f:\mathbb{R}^n\to\mathbb{R}^c$, where  $n\in\mathbb{N}$ represents the dimension of the input space and $c\in\mathbb{N}$ denotes the number of classes.

%We can hence define formally define, without loss of generality, our interpreted model we define the model as $f:\mathbb{R}^n\to\mathbb{R}^c$, where $n\in\mathbb{N}$ where $n\in\mathbb{N}$ represents the input space dimension and $c\in\mathbb{N}$ denotes the number of classes.

Our focus is on \emph{local} explanations --- given an input $\x\in\mathbb{R}^n$, we aim to explain why $f(\x)$ is classified into a particular class: $\hat{y}:=\argmax_j f(\x)_{(j)}$. 
In the \emph{post-hoc} setting, we analyze and explain these local classification decisions based on a given trained model $f$. 
Conversely, in the \emph{self-explaining} setting, we modify $f$ to produce an additional explanatory output and incorporate constraints during training to ensure the generated explanations meet specific desiderata.

%Since our focus deals with explaining NAP tasks, we can generalize this as an explanation for a \emph{classification} task (particularly, the model that we aim to explain $f$ has a few different outputs including regression outputs, however the component which we aim to provide an explanation over is essentialy a classification output). Formally, we define a model $f:\mathbb{R}^n\to\mathbb{R}^c$, where $n\in\mathbb{N}$ is the size of the input space and 
%$c\in\mathbb{N}$ is the number of classes. The explanations we focus on here are \emph{local} explanations, or in other words given some input $\mathbf{x}\in\mathbb{R}^n$, we aim to explain why $f(\x)$ was classified to class: $t:=\argmax_j f(\x)_{(j)}$ . In the \emph{post-hoc} setting we aim to explain these local decisions given some trained model $f$. In the \emph{self-explaining} setting we adjust $f$ to output a second output, and intervene within the training of $f$ so that the generated explanation satisfies some desired criteria.

\subsection{Sufficient Explanations}

A widely studied approach to explaining the decision of a classification model $f$ for a given input instance $\x$ involves identifying a \emph{sufficient} explanation ~\cite{ribeiro2018anchors,carter2019made,bassan2023towards,ignatiev2019abduction}. 
This refers to a subset of input features $S \subseteq \{1, \ldots, n\}$ such that when the features in $ S$ are fixed to their corresponding values in $\x$, the model's classification remains  $\hat{y}:=\argmax_j f(\x)_{(j)}$  with high probability $ \delta $. 
The values for the complementary features in $\overline{S}$ are drawn from a conditional distribution $\mathcal{D}$ over the input space, assuming that the features $S$ are fixed. 
Formally, a sufficient explanation $S$ for a model $f$ and an input $\x$ is defined as:
\begin{equation}
\begin{aligned}
\label{explanation_definition}
\pr_{\z\sim \mathcal{D}} [\argmax_i \ f(\z)_i= \argmax_j \ f(\x)_j \ | \ \z_S=\x_S]\geq \delta
\end{aligned}
\end{equation}

where the expression $\z_S=\x_S$ signifies that we fix the features of the subset $S$ in the vector $\z$ to their corresponding values in $\x$. 

Finally, we observe that choosing the subset $S$ as the entire input space $\{1,\ldots,n\}$ trivially guarantees sufficiency, and in general, larger subsets are more likely to be sufficient. 
Consequently, most studies focus on identifying subsets that are not only sufficient but also of \emph{minimal cardinality}~\cite{ignatiev2019abduction,barcelo2020model,bassan2023towards}, based on the widely held belief that smaller subsets offer better interpretability by containing less information~\cite{ribeiro2018anchors,ignatiev2019abduction}.

\subsection{LSTM}

In this work, the core component of our model utilizes an \acrshort{lstm} architecture.
\acrshort{lstm} networks are a specialized type of \acrfullpl{rnn} designed to overcome the vanishing and exploding gradient problems inherent in traditional \acrshortpl{rnn}~\cite{hochreiter1998vanishing}. 
Initially introduced in 1997~\cite{hochreiter1997long}, \acrshortpl{lstm} are equipped to handle long-term dependencies in sequence data effectively.

An \acrshort{lstm} unit consists of a cell state $ c_t $ and three gates: the input gate $ i_t $, the forget gate $ f_t $, and the output gate $ o_t $. 
The operations of an \acrshort{lstm} cell at time step $t $ can be summarized as follows:

\begin{enumerate}
    \item \textbf{Forget Gate.} Determines which parts of the cell state to retain:
  $$
  f_t = \sigma(W_f \cdot [h_{t-1}, x_t] + b_f)
  $$
  \item \textbf{Input Gate.} Updates the cell state by introducing new information:
  \begin{equation}
  \begin{aligned}
  i_t = \sigma(W_i \cdot [h_{t-1}, x_t] + b_i), \\ \quad \tilde{c}_t = \tanh(W_c \cdot [h_{t-1}, x_t] + b_c)
\\
  c_t = f_t * c_{t-1} + i_t * \tilde{c}_t
  \end{aligned}
  \end{equation}
\item \textbf{Output Gate.} Produces the hidden state $ h_t$ that influences both the output and the next state:
  $$
  o_t = \sigma(W_o \cdot [h_{t-1}, x_t] + b_o), \quad h_t = o_t * \tanh(c_t)
  $$
  \end{enumerate}

\section{Method}
\label{sec:method}

\subsection{An LSTM for NAP}
\label{subsec:lstm4seq}
In this work, we construct our self-explaining architecture on top of a widely used \acrshort{lstm}-based \acrshort{rnn}~\cite{tax2017lstm}. 
In their work, the model’s performance was evaluated on two datasets: BPI12wc and Helpdesk. 
We re-implement the architecture that achieved the best and second-best performance on these datasets, respectively.

\begin{figure}[h]
  \centering
    \includegraphics[height=6cm]{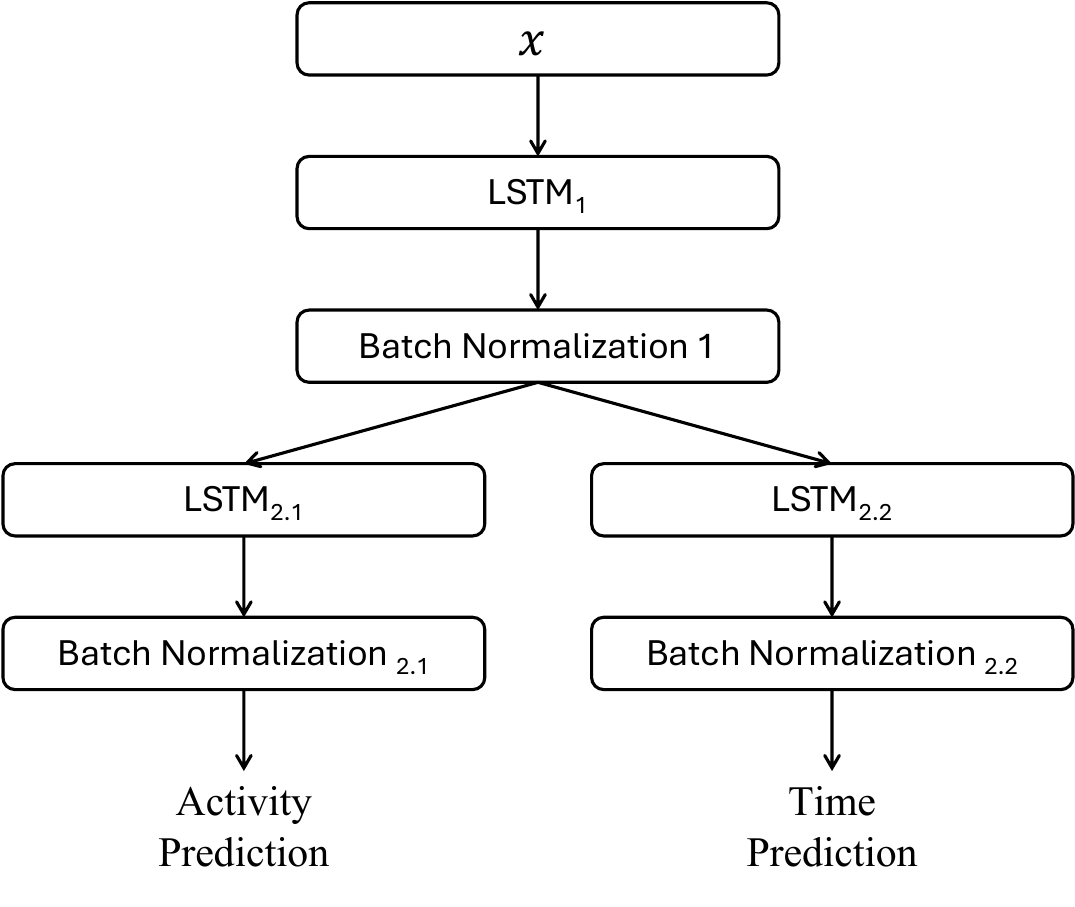}
  \caption{An illustration of the ``traditional'' \acrshort{lstm}-based \acrshort{rnn} architecture~\cite{tax2017lstm}.}
  \label{fig:rnn}
\end{figure}

The \acrshort{rnn} consists of three \acrshort{lstm} layers, with one layer shared between the \acrshort{nap} task and the prediction of the next event's timestamp (see Figure~\ref{fig:rnn}). 
The model takes as input a tensor $x \in \mathbb{R}^{k\times (|A|+m)}$, where $k$ is the maximum number of events in any case within the dataset, $|A|$ represents the number of event types, and $m$ denotes the number of additional features per time point ($m=5$, following Tax et al.; see subsection~\ref{subsec:datasets} for details).

The first (shared) \acrshort{lstm} layer has an input size of $|A|+m$, while the remaining \acrshort{lstm} layers take inputs of size $100$. 
Each \acrshort{lstm} layer contains hidden layers with $100$ units. 
Specifically, the shared \acrshort{lstm} layer consists of two hidden layers, while the other two \acrshort{lstm} layers each have a single hidden layer. 
To mitigate overfitting, all \acrshort{lstm} layers incorporate a dropout rate of $0.2$.

Batch normalization is applied to the output of each preceding \acrshort{lstm} layer. 
The \textit{activity prediction} stream (for \acrshort{nap}) concludes with a Softmax output layer of size $|A|+1$, accommodating \acrfull{eos} prediction. 
Meanwhile, the \textit{time prediction} stream (responsible for predicting the next event's timestamp) employs a simple sum output layer without an activation function.

% \end{wrapfigure}

In the current work we mostly ignore the timestamp prediction but leave it in the architecture for comparability.
While the original \acrshortpl{rnn} was implemented in Keras~\cite{chollet2015keras}, we implemented our models in pyTorch~\cite{paszke2019pytorch}, as our approach required modifications that are not supported by Keras.

The model's training objective simultaneously encompasses both classification (\acrshort{nap}) and regression (predicting the next activity's timestamp). 
Consequently, Tax et al.~\cite{tax2017lstm} formulated the overall joint loss function, $\mathcal{L}_{\theta}$, as follows:

\begin{equation}
    \begin{aligned}
        \mathcal{L}_{\theta} := \mathcal{L}_{CE}(f_{NAP}(\x), a) + \mathcal{L}_{MAE}(f_T(\x), t)
    \end{aligned}
    \label{eq:tax_loss_function}
\end{equation}

%The training objective of the model includes both classification (\acrshort{nap}), and regression (next activity's timestamp) simultaneously.
%Thus, Tax et al.~\cite{tax2017lstm} defined the overall joint loss function $\mathcal{L}_{\theta}$ of the model as follows:
%\begin{equation}
%    \begin{aligned}
%        \mathcal{L}_{\theta} := \mathcal{L}_{CE}(a, \argmax_{j}f_{NAP}(\x)_{(j)}) + \mathcal{L}_{MAE}(t, f_T(\x))
%    \end{aligned}
%    \label{eq:tax_loss_function}
%\end{equation}
 Where $f_T(\x)$ is the timestamp prediction for $\x$, $f_{NAP}$ is the \acrshort{nap}, $a$ is the ground truth actual next activity of $\x$, and $t$ is $a$'s ground truth actual timestamp. 
 $\mathcal{L}_{CE}$ denotes the standard \acrfull{ce} loss and $\mathcal{L}_{MAE}$ denotes the standard mean absolute error loss, or in other words:

\begin{equation}
\begin{aligned}
    \mathcal{L}_{CE}(y,\hat{y}):=-\frac{1}{N}\sum^{N}_{i=1}\sum^{c}_{j=1}y_{i,j}\text{log}\Big(\frac{e^{\hat{y_{i,j}}}}{\sum_{k=1}^c e^{\hat{y_{i,k}}}}\Big),\\ 
    \mathcal{L}_{MAE}(y,\hat{y}):=\frac{1}{N}\sum^{N}_{i=1}||\hat{y_i}-y_i|| \quad \quad \quad
\end{aligned}
\end{equation}

Where $N$ represents the batch size, $\hat{y}$ represents the output vector of either the $f_{NAP}(\x)$ component (in the case of $ \mathcal{L}_{CE}$) or the output of the $f_T(\x)$ component (in the case of $ \mathcal{L}_{MAE}$), and $y$ corresponds to the ground-truth vector, which is a one-hot-encoded representation of the ground-truth target class, or ground-truth timestamp. 
While the original model~\cite{tax2017lstm} employed the \textit{Nadam} optimizer~\cite{dozat2016incorporating} for weight optimization, we utilize the more commonly used \textit{Adam} optimizer~\cite{kingma2014adam}, as it produced superior results. 
Finally, following \cite{tax2017lstm}'s work, we employed a learning rate of $0.002$ for optimization.

%where $y$ is the output of either the $f_{NAP}$ component (in the $\mathcal{L}_{CE}$ case) or the output of $f_T$ (in the $\mathcal{L}_{MAE}$ case), and $\hat{y}$ denotes the corresponding ground-truth vector corresponding to a one-hot-encoded vector over the ground truth class $t$. While~\cite{tax2017lstm} used the \textit{Nadam}~\cite{dozat2016incorporating} learning algorithm to optimize the weights, we use the more standard \textit{Adam} optimizer~\cite{kingma2014adam}, as it yielded better results.
%Finally, following~\cite{tax2017lstm}, we employed a learning rate of $0.002$ for optimization.

\subsection{A Self-Explaining LSTM for NAP} 
In the self-explaining setting, our objective is to enhance our model $f$ by incorporating an explanation component as an additional output. 
Referencing the structure proposed by Tax et al.~\cite{tax2017lstm}, our model $f$ currently includes two outputs: $ f_{NAP}(\x)$, which predicts the next activity, and $f_T(\x)$, which predicts the associated timestamp. 
We extend this architecture to feature three distinct outputs: 
\begin{inparaenum}[(i)] 
\item the \acrshort{nap} (i.e., $f_{NAP}(\x)$), 
\item the timestamp of the next event (i.e., $f_{T}(\x)$), and 
\item the \emph{explanation component}, which elucidates the reasoning behind the specific prediction of the next activity (the explanation for the \acrshort{nap} component), which we denote as $f_E(\x)$.
\end{inparaenum}

%In the self-explaining setting we seek to train our model $f$ with an additional output --- the explanation component. We recall that our model $f$, when based on the framework used by~\cite{tax2017lstm} is already composed of two outputs --- $f_{NAP}$, represeinting the next activity prediction, and $f_T$ represengint the associated timestamp. We hence will adapt our self-explaining architecture to include three overall outputs: \begin{inparaenum}[(i)]
%\item \acrshort{nap}, \item the timestamp of the next event, and \item the explanation component which provides an explanation for the reason the model has obtained this specific next activity prediction (a.k.a, an explanation for the \archshort{NAP} component)\end{inparaenum}.

\begin{figure*}[ht]
  \centering
    \includegraphics[width=15cm]{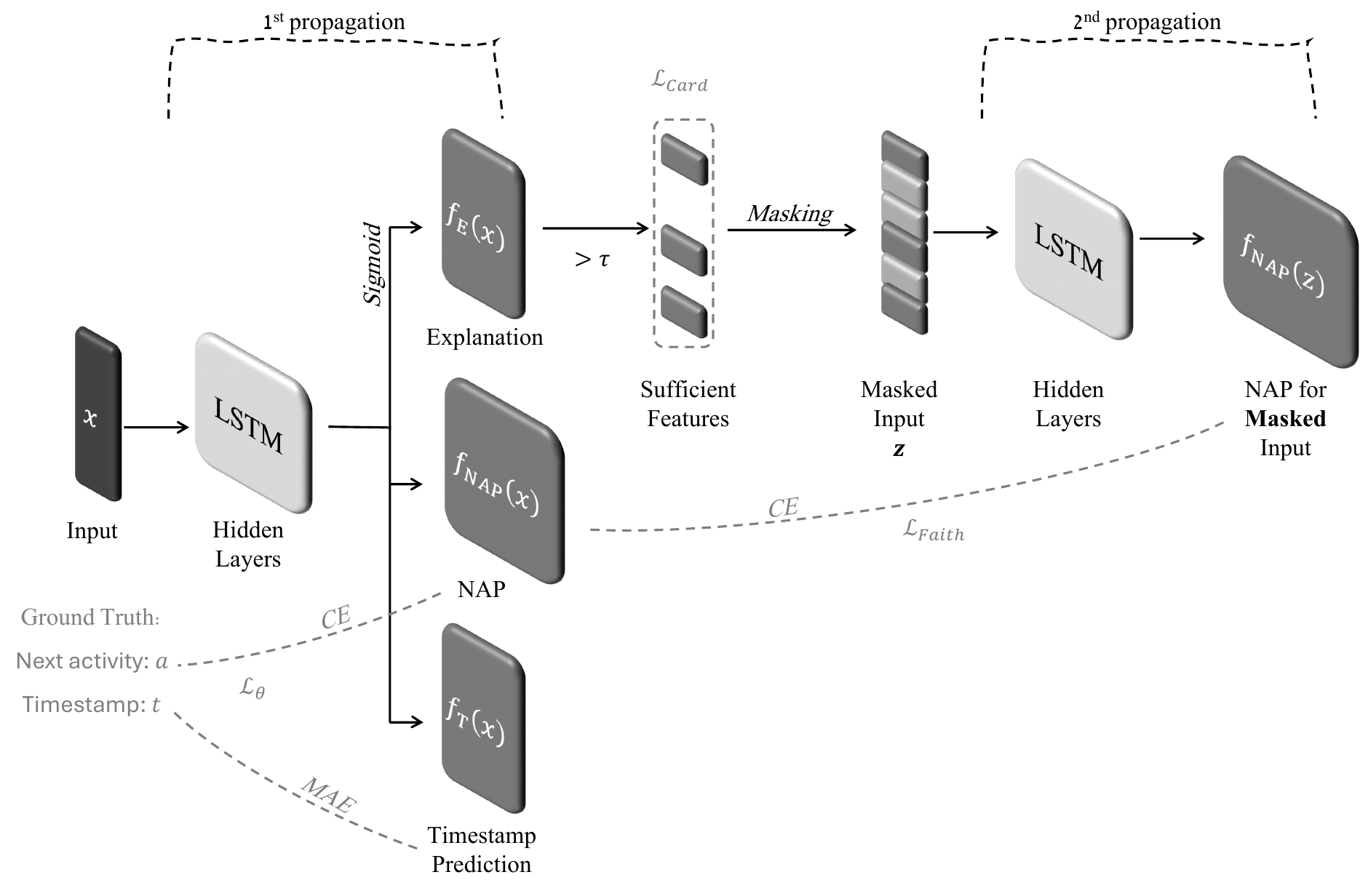}
  \caption{An illustration of the dual propagation procedure used in our self-explaining framework, along with the new loss terms: the faithfulness loss $\mathcal{L}_{Faith}$, which ensures the generated explanation is \emph{sufficient}, and the cardinality loss $\mathcal{L}_{Card}$, which ensures the explanation remains \emph{concise}. It is important to note that the hidden layers of the model are shared across both propagations.}
  \label{fig:architecture}
\end{figure*}

Our approach aligns with the strategy proposed by Bassan et al.~\cite{bassan2025iclr}, which adapts the broader self-explaining framework~\cite{alvarez2018towards} to the domain of sufficient explanations. 
Specifically, rather than conventional propagation through the model during training, we engage in a \emph{dual propagation} process (illustrated in Figure~\ref{fig:architecture}).

In the first propagation, similarly to Tax et al., we optimize the classic prediction components.
In the second propagation, our attention is directed towards optimizing the explanatory component $f_E$. 
This component outputs a tensor of size $k \times (|A|+m)$, which corresponds to the input size, and is transformed via a Sigmoid layer into values between $0$ and $1$. 
We then select all values that surpass a set threshold $\tau$ (for simplicity, $\tau:=0.5$), and deem these as our \emph{sufficient} explanation features, $S$. 
The other features, denoted as $\overline{S}$, are drawn from a distribution $\mathcal{D}$. 
Next, we create a \emph{masked input} $\z$ by maintaining the values in $S$ at their original values in $\x$, while sampling the values in $\overline{S}$ from $\mathcal{D}$. 
This modified input is re-propagated through the model to ensure that the original \acrshort{nap}, $f_{NAP}(\x)$, aligns with the \acrshort{nap} over the masked input, $f_{NAP}(\z)$, thus confirming the sufficiency of $S$. 
Additionally, we introduce a third optimization goal aimed at achieving \emph{minimal cardinality} for the subset $S$.

%During the second propagation, our focus shifts to optimizing the explanatory component $f_E$. The output of $f_E$, a vector of size $n$ representing the number of activities, is processed through a Sigmoid layer to yield values ranging from $0$ to $1$. Subsequently, we identify all values exceeding a predefined threshold $\tau$ (set for simplicity at $\tau:=0.5$), selecting these as our \emph{sufficient} explanation features $S$. The remaining features within $\overline{S}$ are sampled from a specified distribution $\mathcal{D}$. We then construct a \emph{masked input}, which consists of the values in $S$ that were fixed to $\x$ and the remaining values in $\overline{S}$ that were sampled from $\mathcal{D}$. This new masked input is then propogated back through the model with the aim of the original NAP prediction $f_{NAP}(\x)$ over the original input assimilating the prediction of the second NAP prediction over the masked input, and thus validating the sufficiency of the selected subset $S$. We also need to add a third optimization objective which will aim for the selected subset to opt for \emph{minimal cardinality}.

Our updated loss term integrates several components. 
First, it includes the two components from the original model\cite{tax2017lstm}. 
Second, we introduce two new loss terms. 
The first, termed the \emph{faithfulness loss} ($\mathcal{L}_{Faith}$), ensures that the subset $S$ extracted from the output of $f_E$ is \emph{sufficient} with respect to $f_{NAP}$. 
This is achieved by employing the standard \acrshort{ce} loss $\mathcal{L}_{CE}$ to minimize the difference between the predicted probabilities from the first propagation, $f_{NAP}(\x)$, and those of the masked input, $f_{NAP}(\z)$. 
The final loss term aims to encourage the sufficient subset $S$ to be compact, optimized through the standard $\mathcal{L}_1$ loss, which promotes sparsity in the explanation component $f_E(\x)$. 
Therefore, we can formalize our final and overall loss term as follows:

\begin{equation}
    \begin{aligned}
        \mathcal{L}_{\theta} := \mathcal{L}_{CE}(f_{NAP}(\x), a) + \mathcal{L}_{MAE}(f_T(\x), t)+\\ \lambda \mathcal{L}_{faith} +
        \xi \mathcal{L}_{Card} \quad \quad \quad \quad
    \end{aligned}
    \label{eq:loss_function}
\end{equation}

where we have that:
\begin{equation}
    \begin{aligned}
        \mathcal{L}_{Faith} := \mathcal{L}_{CE}(f_{NAP}(\z),\argmax_j f_{NAP}(\x)_{(j)}), \\
        \mathcal{L}_{Card} := \mathcal{L}_1(f_E(\x))=||f_E(\x)||_1 \quad \quad  
    \end{aligned}
    \label{eq:loss_help}
\end{equation}

And $\z$ represents the masked input obtained during the second propagation phase, where we select the subset $S:=\{i \ | \ f_E(\x)_i\geq\tau\}$ and replace the remaining features in $\overline{S}$ by sampling from the conditional distribution $\mathcal{D}(\z|\z_S=\x_S)$.

\section{Experiments}

\label{sec:experiments}

\subsection{Benchmarks}

\label{subsec:datasets}

Table \ref{table:datasets} provides descriptions of four publicly available datasets used for evaluating our approach. 
These datasets consist of real-world event logs from \acrshort{it} support and the banking industry. 
All datasets are accessible through the 4TU Center for Research Data~\footnote{https://data.4tu.nl/info/en/}.
\begin{table*}[h]
\caption{Description of the benchmarks used for evaluation}
\label{table:datasets}
\centering
\begin{tabular}{lcccccc}
\toprule
%
% Dataset & Num.   & Num.    & Num.       & Avg.\ case & Max case  \\
%         & cases  & events & activities  & length     &  length   \\
Dataset & \#Cases  & \#Events & \#Activities  & Avg. case length & Max case length   \\
%\\[0.3cm]
%
\midrule
%\hline
Helpdesk  & 4580   & 21348  & 14  & 4.66  & 15  \\
BPI12wc   & 9658   & 72413  & 6   & 7.50  & 74  \\
BPI13in   & 7554   & 65533  & 13  & 8.68  & 123 \\
BPI17o    & 42995  & 193849 & 8   & 4.51  & 5 \\ 
\bottomrule
\end{tabular} 

\end{table*}

% Datasets description
The Helpdesk dataset comprises of event logs from a ticket management process within an Italian software company. 
BPI13in describes incident logs from Volvo \acrshort{it}.
BPI12wc is a subset of the popular BPI12 dataset that contains logs from loan application process instances within a Dutch financial institution. 
These instances commence with a customer submitting a loan application and conclude with a decision on the application: approval, cancellation, or rejection.
% Given the popularity of the BPI12 dataset in \acrshort{pbpm} research, an expanded dataset, BPI17, was collected from the same financial institution five years later
% BPI17o is a subset of BPI17 that comprises of events corresponding to the states of loan offers communicated to customers.
BPI17 is an expanded version of BPI12 which was collected in the same financial institution five years later.
BPI17o is a subset of BPI17 that comprises of events corresponding to the states of loan offers communicated to customers.

% Notice that \acrshort{it} support and the banking industry, are important domains for explainability applications. 
% For example, consider the \acrshort{nap} problem and assume that our algorithm predicted a ``Refused'' state, indicating that the customer would reject the loan conditions offered by the bank.
% In this case, providing a sufficient explanation for this prediction can hold practical significance. 
% It can aid in comprehending the loan application process and uncovering potential issues therein.

% Feature description
In our work, we used the same set of features used by Tax et al.~\cite{tax2017lstm}: \begin{inparaenum}[(i)] \item activity type in one-hot-encoding, \item event index ($i \geq 1$) in the process trace, \item time since the first event in the process trace, \label{try_this}\item time since the previous event in process trace, \item time since midnight, and \item numeric weekday\end{inparaenum}.
Features \RNum{3} and \RNum{4} were divided by their respective means in the training data.
Features \RNum{5} and \RNum{6} were normalized within the $[0,1]$ scale.

\subsection{Experimental Setup}
\label{subsec:LSTM}
We implemented and evaluated two \acrshort{lstm} \acrshort{rnn} architectures on all the datasets presented in Table \ref{table:datasets}, which included the ``traditional'' LSTM model~\cite{tax2017lstm} and our self-explaining model, that modifies this architecture. 

We adapted the method described in section \ref{sec:method} for these datasets. 
As outlined in subsection~\ref{subsec:lstm4seq}, the \acrshort{nap} output integrates the dataset's activity types and an additional activity corresponding to \acrshort{eos}. 
Given the sequential nature of \acrshort{bpm} data, where events should appear in chronological order, we established that an event sequence where event $k+1$ precedes event $k$ (with $0<k\in\mathbb{R}$) would not represent a valid process trace. 
Consequently, we fixed the event index features as an inherent part of the sufficient explanation, thus excluding them from the explanation output.

Considering the varied lengths of cases, those shorter than the maximum case length were padded with zeros to the left of the event sequence. 
This introduces a challenge: altering the feature values of these ``dummy'' events could be problematic, conceptually. 
Alternatively, designating all features of these events as fixed would categorize them as part of the sufficient subset, inadvertently expanding the size of these subsets beyond the model's control. 
Additionally, such features would not provide meaningful explanations as they stand. 
In this study, we chose the former approach and refined the explanations to omit features from any ``dummy'' events. 
It should be noted that this adjustment only affects the visual representations in Figures~\ref{fig:explanations_ex1} and~\ref{fig:explanations_ex2}, whereas all explanation-related metrics, like the average size of explanations, included these features.

The instances in each dataset were organized in chronological order based on the timestamp of their first events. 
The first two-thirds of these instances were allocated for training and validation (with a 90\%-10\% split), and the remaining third was designated for testing.

\subsection{Grid Search}
\label{subsec:hyperparameters}

% \begin{wraptable}{L}{80mm}
% % \begin{table}[h!]
% \caption{Optimal hyperparameters per dataset}
% \label{table:hyperparameters}
% \centering
% \begin{tabular}{lcc|cc}
% %
% \toprule
% %
% & \multicolumn{2}{c}{Entire space} & \multicolumn{2}{|c}{Truncated space} \\
% \cmidrule(lr){2-3}\cmidrule(lr){4-5}
% Dataset & Learning rate & $\xi$ & Learning rate & small $\xi$ \\ 
% %\\[0.3cm]
% %
% \midrule
% %\hline
% Helpdesk  &  0.01  & $10^{-8}$ &  0.0001 & $10^{-9}$ \\
% BPI12wc   &  0.001 & $10^{-9}$ &  0.001  & $10^{-9}$ \\
% BPI13in   &  0.001 & $10^{-5}$ &  0.001  & $10^{-9}$ \\
% BPI17o    &  0.001 & $10^{-7}$ &  0.001  & $10^{-10}$ \\
% % 
% \bottomrule
% \end{tabular} 

% % \end{table}
% \end{wraptable}
%
% hyperparameters
We conducted hyperparameter optimization within a configuration space of $30$ combinations, focusing on two key parameters essential for the algorithm's convergence: the learning rate and the cardinality loss coefficient, $\xi$. 
We utilized a grid search approach for the learning rate within $\{10^{-2}, 10^{-3}, 10^{-4}, 10^{-5}, 10^{-6}\}$ and for $\xi$ within $\{10^{-5}, 10^{-6}, 10^{-7}, 10^{-8}, 10^{-9}, 10^{-10}\}$. 
Lower values of $\xi$ lessen the influence of the cardinality component in the loss function, leading to larger explanations and fewer empty explanations, which may increase the proportion of sufficient explanations. 
Consequently, we also explored a narrowed hyperparameter space with $\xi \in \{10^{-9}, 10^{-10}\}$. 
Below, we will explore the trade-offs involved in these hyperparameter selection strategies. 
The best-performing combinations of learning rate and $\xi$ on the validation sets were then applied for the final assessments on the test sets. 
We fixed the faithfulness loss coefficient $\lambda$ at $1$ in the loss function (Equation~\ref{eq:loss_function}) and set the feature selection threshold $\tau$ at the default value of $0.5$.

\subsection{Evaluation Metrics}

Building on previous work on sufficient explanations~\cite{ignatiev2019abduction,bassan2023towards,wu2024verix}, we assess the quality of explanations --- whether derived from traditionally trained models using post-hoc methods or our self-explaining approach --- using the following metrics: \begin{inparaenum}[(i)]  
    \item mean explanation \emph{size}, favoring smaller explanations,  
    \item \emph{faithfulness}, measured as the proportion of explanations verified as sufficient, assessed by sampling the complement of the generated subset from a uniform distribution, and  
    \item the mean \emph{computation time} required to generate the explanations.  
\end{inparaenum}

\subsection{Results}
\label{subsec:results}
 
\subsubsection{Do the self-explaining trained models have a reduced performance?}
Evaluating the potential performance decrease that our self-explaining approach may entail compared to standard training is one of the critical practical considerations in our work.
% We used the prediction accuracy as the evaluation metric for comparing different prediction approaches. 
We used the prediction accuracy to evaluate performance. 
Consistent with Tax et al.~\cite{tax2017lstm}, accuracy was computed for all process traces and subtraces of length $>1$. 
% A dummy activity labeled ``end of trace'' was included in the list of feasible next activities.
 
Table~\ref{table:accuracy} presents a comparison of accuracies on the testing sets between the re-implementation of Tax et al.'s work~\cite{tax2017lstm} and our approach, utilizing the hyperparameter selection considering all $\xi$ values or only small $\xi$ values.
Our results suggest that, on average, the performance of our approach does not perform substantially worse than the re-implementation, and in certain cases it even outperforms it. 
Furthermore, truncating the hyperparameter space (column ``small $\xi$'') does not appear to result in a significant performance drop as compared to the entire space, with the exception of the Helpdesk dataset.

% \begin{wraptable}{R}{70mm}
\begin{table}[h!]
\caption{Summary of model accuracies}
\label{table:accuracy}
\centering

\begin{tabular}{lc|cc}
\toprule
& Tax et al. & & \\
& ~\cite{tax2017lstm} & \multicolumn{2}{|c}{Our approach} \\
Dataset & approach  & $\xi$ & small $\xi$ \\
%\\[0.3cm]
%
\midrule
%\hline
Helpdesk  &  0.669  & 0.799 & 0.730 \\
BPI12wc   &  0.779  & 0.771 & 0.771 \\
BPI13in   &  0.692  & 0.709 & 0.689 \\
BPI17o    &  0.755  & 0.718 & 0.714 \\ 
\bottomrule
\end{tabular} 

\end{table} 
% \end{wraptable}

\subsubsection{Comparison with Anchors}

We compared the explanations produced by our self-explaining approach with those generated by the well-known post-hoc \acrshort{xai} method, Anchors~\cite{ribeiro2018anchors}, which also provides sufficient explanations for standard trained models. 
Due to the high computational cost of Anchors, we limited the comparison to \acrshortpl{nap} for the first $200$ case instances in the testing sets.
%
% \begin{table}[h!]
% \caption{Anchors approach: Existing and sufficient explanations}
% \label{table:anchors_explanations}
% \centering
% %
% \begin{tabular}{lccc}
% %
% \toprule
% %
% Dataset & Existing & Sufficient   & Sufficient  \\
% & explanations, \%  & explanations  & explanations  \\ &                  & out or existing, \%  & overall, \% \\
% %\\[0.3cm]
% %
% \midrule
% %\hline
% Helpdesk  &  95.26 & 16.52 & 15.74  \\
% BPI12wc   &  25.60 & 16.22 & 4.15  \\
% BPI13in   &  4.26  & 52.17 & 2.22 \\
% BPI17o    &  100.0 & 8.78  & 8.78  \\ 
% % 
% \bottomrule
% \end{tabular} 

% \end{table}

\begin{table*}[h!]
\caption{The number of cases resolved within the timeout ($600$ seconds) for which explanations were provided (existing explanations) and the faithfulness score, represented by the percentage of cases verified as sufficient, for both our self-explaining approach and for explanations obtained via Anchors on standard trained models}
\label{table:explanations_stats}
\centering
\begin{tabular}{lccc|cc}
\toprule
& \multicolumn{3}{c}{Anchors} & \multicolumn{2}{|c}{Our approach} \\
\cmidrule(lr){2-4}\cmidrule(lr){5-6}
& Existing & Sufficient & Sufficient & \multicolumn{2}{|c}{Sufficient}  \\
& explanations, \% & explanations out & explanations & \multicolumn{2}{|c}{explanations, \%}  \\
Dataset & & of existing, \% & overall, \% & $\xi$ & small $\xi$ \\

% Dataset & Existing & Sufficient & Sufficient  Existing & \multicolumn{2}{c}{Sufficient}  \\
% & explanations & \multicolumn{2}{c}{explanations}  & explanations  & \multicolumn{2}{c}{explanations}  \\
% & explanations & explanations & explanations & out or existing  & overall & Tables 2 and 3 & Table 2 & Table 3 \\
%\\[0.3cm]
%
\midrule
%\hline
Helpdesk  &  95.26 & 16.52 & 15.74 & 41.36 & 73.68   \\
BPI12wc   &  25.60 & 16.22 & 4.15  & 63.54 & 63.54  \\
BPI13in   &  4.26  & 52.17 & 2.22  & 48.01 & 60.59 \\
BPI17o    &  100.00 & 8.78  & 8.78  & 67.35 & 80.80  \\ 

\bottomrule
\end{tabular} 

\end{table*}

Table \ref{table:explanations_stats} summarizes the results of the explainability experiments using the Anchors method. We ran it with a $600$ seconds timeout. 
As a result, the column showing the percentage of existing explanations reflects the proportion of predictions for which an explanation was not identified within the $600$ second timeframe.

%Table \ref{table:explanations_stats} provides a summary of explainability experiments for the Anchors approach.  
%If the algorithm failed to find an explanation within $600$ seconds, we terminated the process.  
%Consequently, the column displaying the percentage of existing explanations indicates the fraction of predictions where an explanation could not be found within $600$ seconds.

We observed that Anchors explanations were absent for most data points in the BPI12wc and BPI13in datasets within the $600$-second timeframe, as shown in Table \ref{table:datasets}. 
These datasets exhibit notably longer maximum trace lengths and, consequently, a greater number of model features compared to the Helpdesk and BPI17o datasets. 
Across all datasets, the proportion of Anchors explanations verified as sufficient (i.e., the explanation's faithfulness) was generally low (under $20\%$ for all benchmarks).

%
% \begin{table}[h!]
% \caption{our approach: Existing and sufficient explanations}
% \label{table:our_explanations}
% \centering

% \begin{tabular}{lccc}
% %
% \toprule
% %
% Dataset & Existing  & \multicolumn{2}{c}{Sufficient}  \\
% & explanations, \%  & \multicolumn{2}{c}{explanations, \%}  \\
% %
% %\midrule
% %
% %Hyperparameters
% %
% & Tables \ref{table:hyperparameters} and \ref{table:hyperparameters_small_xi} & Table \ref{table:hyperparameters} & Table \ref{table:hyperparameters_small_xi} \\ %\\[0.3cm]
% %
% \midrule
% %\hline
% Helpdesk  & 100.0 & 41.36 & 73.68  \\
% BPI12wc   & 100.0 & 63.54 & 63.54 \\
% BPI13in   & 100.0 & 48.01 & 60.59 \\
% BPI17o    & 100.0 & 67.35 & 80.80  \\ 
% % 
% \bottomrule
% \end{tabular} 
% \end{table}

Table \ref{table:explanations_stats} displays the proportions of existing and sufficient explanations for both Anchors and our method. 
Our approach consistently yields explanations, with a higher percentage of these explanations being verified as sufficient (i.e., greater faithfulness) compared to Anchors across all datasets and methods of hyperparameter selection. 
For instance, for the dataset BPI12wc only roughly one in four data points are explainable by Anchors in a timely manner.
Out of those explanations, only roughly one in six is sufficient, resulting in only roughly one of $24$ data points being sufficiently explained by Anchors in a timely manner.
In contrast, our method, for the same dataset, is able to sufficiently explain roughly five to six out of ten data points in a timely manner.
Additionally, concentrating on smaller $\xi$ values significantly boosts the percentage of sufficient explanations relative to a wider hyperparameter space, in some cases (e.g., Helpdesk) almost doubling this percentage.

%Table \ref{table:explanations_stats} presents the fractions of existing and sufficient explanations for Anchors and for our approach. 
%Notably, explanations are consistently found using our approach, and the percentages of explanations that are verified as sufficient (i.e., the faithfulness), exceed those of Anchors for all datasets and both approaches to hyperparameter selection. 
%Moreover, focusing on small $\xi$ values entails a substantial increase in the percentage of sufficient explanations compared to the broader hyperparameter space. 
%
% \begin{table}[h]
% \caption{Mean size of explanations}
% \label{table:explanation_size}
% \centering
% %
% \begin{tabular}{lccc}
% %
% \toprule
% %
% & \textbf{Anchors} &  \multicolumn{2}{c}{\textbf{Our approach}} \\
% %
% \cmidrule(lr){3-4}
% %Hyperparameters 
% %
% \textbf{Dataset} & & $\xi$ & small $\xi$ \\
% %
% %\\[0.3cm]
% %
% \midrule
% %\hline
% Helpdesk  & 30.78  & 16.08  & 22.61    \\
% BPI12wc   & 78.73  & 102.97 & 102.97   \\
% BPI13in   & 126.85 & 125.05 & 164.08   \\
% BPI17o    & 14.26  &  6.20  & 7.71  \\ 
% % 
% \bottomrule
% \end{tabular}

% \end{table}

Table \ref{table:expalantion_kpis} displays the average size of explanations for Anchors and our algorithm's two hyperparameter settings. 
The size of an explanation is determined by the count of feature values it contains. 
The explanation lengths produced by both methods are similar. 
It is crucial to note that our algorithm inherently counts the feature numbers corresponding to event indices in the size of the explanation. 
This results in the observed larger average explanation sizes for the BPI12wc and BPI13in datasets.

%Table \ref{table:expalantion_kpis} presents the mean size of explanations for Anchors and the two hyperparameter combinations of our algorithm. 
%The size of an explanation is defined as the number of feature values in the explanation. 
%Both approaches yield comparable explanation lengths. 
%It is important to reiterate that our algorithm automatically includes the number of features corresponding to event indices in the explanation's size. 
%This accounts for the larger mean size of explanations observed for the BPI12wc and BPI13in datasets.

\begin{table*}[h]
\caption{Summary of mean explanation sizes and mean computation times for our self-explaining approach compared to explanations produced by Anchors over standard trained models}
\label{table:expalantion_kpis}
\centering

\begin{tabular}{lc|cc|c|cc}
\toprule
 & \multicolumn{3}{c|}{Mean explanation size} & \multicolumn{3}{c}{Mean computation time, sec} \\
 \cmidrule(lr){2-4}\cmidrule(lr){5-7}
 & Anchors & \multicolumn{2}{|c|}{Our approach} & Anchors & \multicolumn{2}{|c}{Our approach} \\
Dataset & & $\xi$ & small $\xi$ & & $\xi$ & small $\xi$ \\
\midrule
%\hline
Helpdesk  & 30.78  & 16.08  & 22.61  & 80.57   &  0.00081 & 0.00081    \\
BPI12wc   & 78.73  & 102.97 & 102.97 & 290.79  &  0.00352 & 0.00352   \\
BPI13in   & 126.85 & 125.05 & 164.08 & 31.10   &  0.00526 & 0.00496   \\
BPI17o    & 14.26  &  6.20  & 7.71   & 17.96   &  0.00040 & 0.00039  \\ 
\bottomrule
\end{tabular}

\end{table*}

% \begin{table}[h]
% \caption{Summary of mean computation times, sec}
% \label{table:response_time}
% \centering
% %
% % \begin{tabular}{lcc}
% % %
% % \toprule
% % %
% % Dataset & Anchors & ours   \\
 
% % %\\[0.3cm]
% % %
% % \midrule
% % %\hline
% % Helpdesk  & 80.57   &  0.00081 \\
% % BPI12wc   & 290.79  &  0.00352 \\
% % BPI13in   & 31.10   &  0.00526  \\
% % BPI17o    & 17.96   &  0.00040 \\ 
% % % 
% % \bottomrule
% % \end{tabular} 

% \begin{tabular}{lccc}
% %
% \toprule
% %
% Dataset & Anchors & Our approach ($\xi$) & Our approach (small $\xi$) \\
% \midrule
% %\hline
% Helpdesk  & 80.57   &  0.00081 & 0.00081    \\
% BPI12wc   & 290.79  &  0.00352 & 0.00352   \\
% BPI13in   & 31.10   &  0.00526 & 0.00496   \\
% BPI17o    & 17.96   &  0.00040 & 0.00039  \\ 
% % 
% \bottomrule
% \end{tabular}

% \end{table}

The right side of Table \ref{table:expalantion_kpis} shows the average computation times per explanation for both our method and Anchors. 
As expected, our method is markedly more efficient, outperforming Anchors by several orders of magnitude. 
Due to its lengthy computation times, Anchors proves impractical for many applications across our analyzed domains.

%The right side of Table \ref{table:expalantion_kpis} displays the mean computation times per explanation for our method and for Anchors. 
%Not surprisingly, it is evident that our approach is significantly more efficient, by several orders of magnitude. 
%The computation times associated with Anchors render it infeasible for many domains in practical applications.

\subsubsection{Explanation examples for our approach and Anchors}

Here we present two instances from the test subset of the BPI12wc dataset, where sufficient explanations were given by both the Anchors method and our self-explaining technique. 
For this dataset, we observed no disparity whether all possible $\xi$ values or only smaller $\xi$ values were considered. 
In both instances, the model predictions were accurate.

%Here we provide two examples from the test set of the BPI12wc dataset, where sufficient explanations were provided by both Anchors and our approach.
%For this dataset, there was no difference between considering all possible $\xi$ values and considering only small $\xi$ values.
%In both examples the predictions of the models were correct.
%
\begin{figure*}[ht!]
  \centering
    \includegraphics[width=13cm]{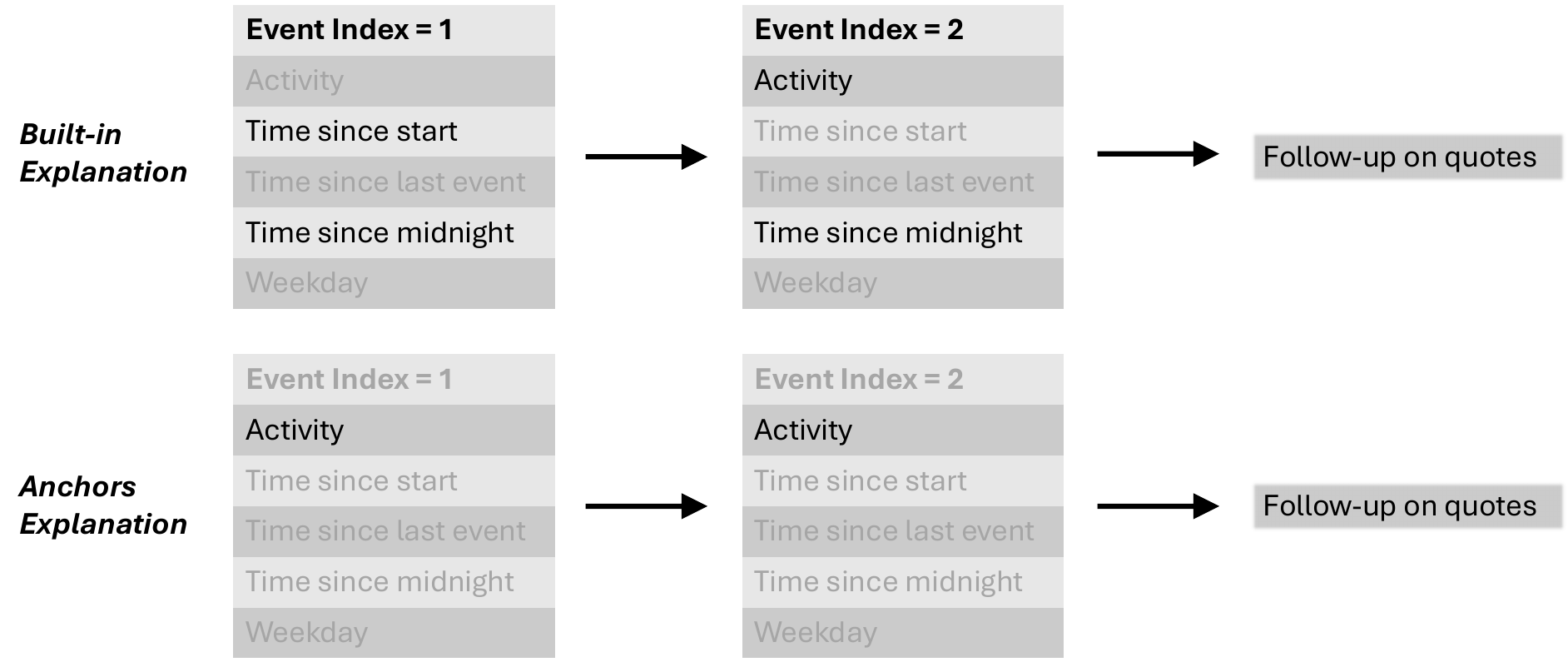}
  \caption{An example of an explanation generated either by Anchors on standard trained models or inherently by our self-explaining approach, when applied to the prediction of the \emph{third} activity in a BPI12wc process. The text in black represents the sufficient explanation, while the features not included in the explanation appear in gray.}
  \label{fig:explanations_ex1}
\end{figure*}

Figure \ref{fig:explanations_ex1} demonstrates the sufficient explanations for the third event within a process instance. 
We note that event indices are inherently included in the explanations through our method. 
Furthermore, our algorithm integrates various time-related features, whereas Anchors incorporates the initial activity of the process instance.

%Figure \ref{fig:explanations_ex1} illustrates the sufficient explanations for the third event in a process instance. 
%We observe that event indices are automatically incorporated into the explanations for our approach. 
%In addition, our algorithm includes several time features, while Anchors adds the first activity of the process instance.  

\begin{figure*}[ht!]
  \centering
    \includegraphics[width=15.6cm]{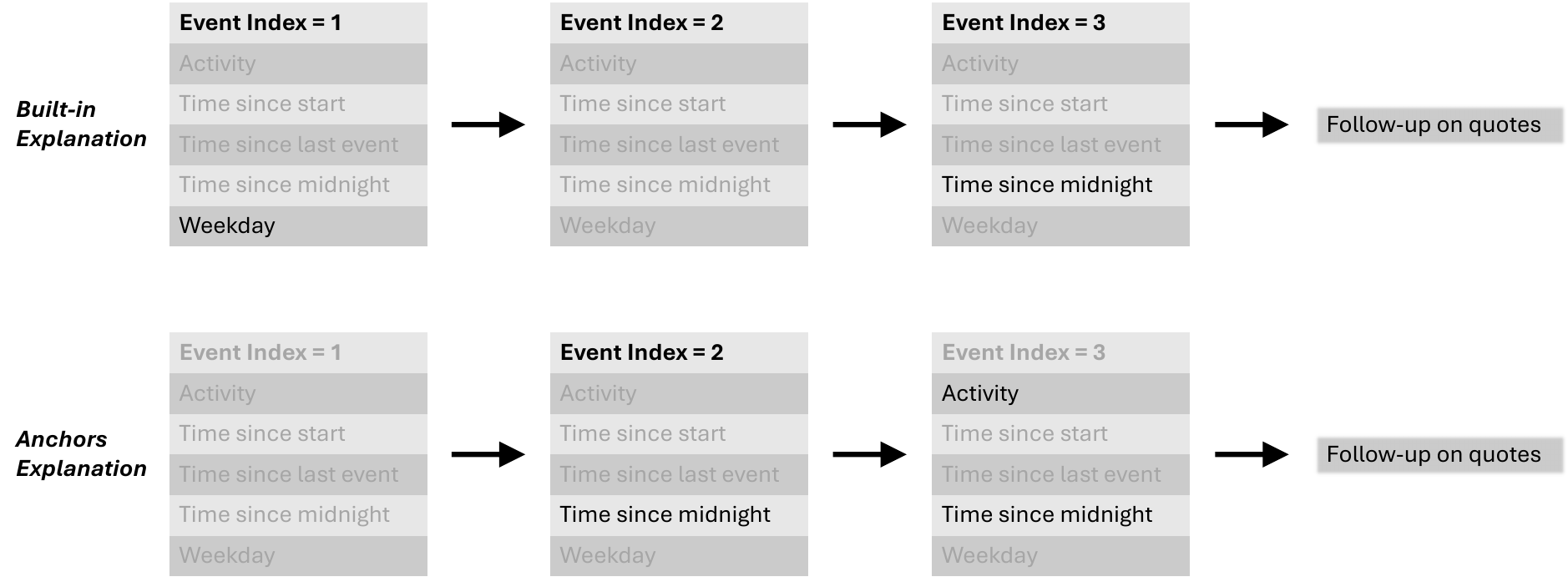}
  \caption{An example of an explanation generated either by Anchors on standard trained models or inherently by our self-explaining approach, when applied to the prediction of the \emph{fourth} activity in a BPI12wc process. The text in black represents the sufficient explanation, while the features not included in the explanation appear in gray.}
  \label{fig:explanations_ex2}
\end{figure*}

Figure \ref{fig:explanations_ex2} displays the explanation for the fourth activity in another instance. 
In this case, both algorithms incorporate some time features into the set of sufficient explanations. 
% Notably, Anchors' explanation includes one of the event indices, although this algorithm cannot accurately handle these features: it is inappropriate to add noise to the event indices to assess explanation sufficiency.
Notably, Anchors' explanation includes one of the event indices, suggesting that the other event indices can be altered and the prediction will remain unchanged. 
We, however, postulate in the current work that modifying the event indices in the input affects the integrity of the input as a process trace.

The explanations provided by both methods for the two cases appear reasonable and share some common features: Activity type of the previous event in Figure \ref{fig:explanations_ex1} and 'Time since midnight' of the previous event in Figure \ref{fig:explanations_ex2}. 
However, it is important to note that we selected examples where Anchors offered valid explanations. 
As indicated in Table \ref{table:explanations_stats}, such examples constitute a relatively small fraction (one in $24$) of the instances in the test set.

\section{Discussion and Future Work}

\label{sec:discussion}

Our experiments on real-world \acrshort{bpm} logs provided several key insights.
First, we observed that incorporating the self-explaining approach does not significantly degrade prediction performance compared to a standard model trained with the same architecture but without the self-explaining component.
Second, our method outperforms a state-of-the-art post-hoc explainability approach applied to a standard-trained model across multiple evaluation metrics.
Additionally, our approach consistently produces explanations with significantly greater efficiency and a much higher faithfulness score --- meaning a larger proportion of explanations are verified as sufficient.

%Our experiments conducted on real-world \acrshort{bpm} logs yielded several important insights. 
%First, we found that the self-explaining approach we have used does not result in a significant deterioration of prediction performance compared to a standard trained model using the same architecture, without the self-explaining component.
%Second, our method surpasses a parallel state-of-the-art post-hoc explainability approach used over a standard-trained model across several comparison metrics. 
%Our approach always generates explanations substantially more efficiently, with a much higher faithfulness score (i.e., a higher percentage of explanations that were indeed verified to be sufficient). 

Additionally, we investigated the trade-off between greater and smaller values of the cardinality coefficient $\xi$ in the loss function. 
By confining the range to smaller values exclusively, we achieve a greater proportion of sufficient explanations, albeit at the expense of potential moderate performance decline.

The present study has several potential limitations. 
\begin{inparaenum}[(i)]
\item While the datasets used are real-world datasets, they are limited in number and relatively small. 
However, the proposed approach is applicable to larger event log datasets, making this an interesting direction for future research. 
\item The effectiveness of explanations in the \acrshort{pbpm} context largely depends on their usefulness to business analysts. 
Thus, beyond the extensive faithfulness evaluations conducted, a human user study could provide valuable insights. 
\item Due to the padding of shorter log traces, their explanations may include ``dummy'' features, though these are removed before presenting the final explanations.
\end{inparaenum}

%The present study has several potential limitations. \begin{inparaenum}[(i)] \item First, although the datasets utilized are real-world datasets, they are limited in number and relatively small. However, the approach can be used on larger sets of event logs, which presents intriguing future work.
%\item Second, the effectiveness of  explanation in the \acrshort{pbpm} context is largely dependent on its value to business analysts. Therefore, in addition to the extensive fatithfulness evaluations we conducted, a human user study can be impactful. \item Lastly, due to the padding of shorter log traces, their explanations are more likely to incorporate dummy (i.e., redundant) features, though these are removed before presenting the explanations.
%\end{inparaenum}

This research can also be continued in several interesting directions. 
Additional \acrshort{pbpm} tasks, such as predicting time until instance completion, suffix, and process outcome, can be addressed using our approach. 
Our methodology can be potentially incorporated into advanced \acrshort{dl} prediction architectures, such as transformers, enabling the extension of prediction models to incorporate additional features that offer contextual insights into the process. 
One can also further explore the trade-off between the three terms of the loss function, corresponding to prediction, faithfulness, and cardinality. 
Lastly, it would be worthwhile to explore more flexible approaches to padding, aiming to reduce input dimensionality for event logs containing lengthy event sequences.

\section{\uppercase{Acknowledgements}}
\label{sec:acknowledgements}

The research leading to the results presented in this paper has received funding from the European Union's funded Project AI4Gov under grant agreement no 101094905.

\bibliographystyle{apalike}
{\small
\bibliography{citations}}

% \section*{\uppercase{Appendix}}

% If any, the appendix should appear directly after the
% references without numbering, and not on a new page. To do so please use the following command:
% \textit{$\backslash$section*\{APPENDIX\}}

\end{document}